\documentclass[conference]{IEEEtran}\IEEEoverridecommandlockouts
\usepackage{hyperref}
\usepackage{bm}
\usepackage{mathrsfs}
\usepackage{amsmath}
\usepackage{amsthm}

\usepackage{paralist}
\usepackage{multirow}
\usepackage{graphicx}
\usepackage{amsfonts}
\usepackage{amsmath,epsfig}
\usepackage{color,xcolor}
\usepackage{soul}
\usepackage{mathtools}
\usepackage{stfloats}
\usepackage{amssymb}
\usepackage{epstopdf}
\usepackage{latexsym}
\usepackage{color}
\usepackage{cite}
\usepackage{algorithm}
\usepackage{algorithmic}
\usepackage[algo2e]{algorithm2e}
\usepackage{float}
\usepackage{subfig}

\ifCLASSINFOpdf

\else

\fi

\begin{document}
%
\title{Device Scheduling for Relay-assisted Over-the-Air Aggregation in Federated Learning}

\author{Fan Zhang, Jining Chen, Kunlun Wang, and Wen Chen

\thanks{F. Zhang, and K. Wang are with the Shanghai Key Laboratory of Multidimensional Information Processing, East China Normal University, Shanghai 200241, China, and also with the School of Communication and Electronic Engineering, East China Normal University, Shanghai 200241, China (e-mail: 51215904110@stu.ecnu.edu.cn; klwang@cee.ecnu.edu.cn).}
\thanks{J. Chen is with the Joint Innovation Laboratory of Digital GuangXi Smart Infrastructures, Guangxi Information Center, Nanning 530000, China (e-mail: chenjn@gxi.gov.cn).}
\thanks{W. Chen is with the Department of Electronic
Engineering, Shanghai Jiao Tong University, Shanghai 200240, China. (email: wenchen@sjtu.edu.cn).}

}
\maketitle

\IEEEpeerreviewmaketitle
\begin{abstract}
 Federated learning (FL) leverages data distributed at the edge of the network to enable intelligent applications. The efficiency of FL can be improved by using over-the-air computation (AirComp) technology in the process of gradient aggregation. In this paper, we propose a relay-assisted large-scale FL framework, and investigate the device scheduling problem in relay-assisted FL systems under the constraints of power consumption and mean squared error (MSE). we formulate a joint device scheduling, and power allocation problem to maximize the number of scheduled devices. We solve the resultant non-convex optimization problem  by transforming the optimization problem into multiple sparse optimization problems. By the proposed device scheduling algorithm, these sparse sub-problems are solved and the maximum number of federated learning edge devices is obtained. The simulation results demonstrate the effectiveness of the proposed scheme as compared with other benchmark schemes.
\end{abstract}

\section{Introduction}

With the development of intelligent Internet of Things (IoT), the number of IoT devices such as mobile phones, smart wearables, and autonomous vehicles will explode. These IoT devices generate huge amounts of data by interacting with the physical environment and the users around them \cite{01}. Data is a key factor driving the development of artificial intelligence \cite{02}. Traditional cloud computing requires devices to transmit data to the cloud data center for centralized processing, and the upload of a large amount of data will cause data privacy leakage and communication resources shortage. With the improvement of computing capabilities of edge devices, \cite{03} proposes a distributed machine learning framework called Federated Learning (FL), which enables multiple edge devices to train a model collaboratively using their own computing resources and data. Because training data is always stored locally, FL can alleviate communication resources and data security problems to a certain extent.

In FL, multiple edge devices train a shared global model using local data sets. When increasing the number of devices involved in the FL process, larger and more diverse data sources can be accessed. The diversity of data helps models understand patterns in data from different perspectives, ultimately developing more accurate models\cite{04}. However, due to channel fading, limited broadband resources, and limited battery capacity of edge devices, the integration of large-scale IoT devices into FL is still a challenge.

In order to achieve large-scale FL, \cite{05} and \cite{06} reduce the amount of data transmitted in the uplink and downlink of FL by data compression techniques, but the distortion caused by compression will reduce the accuracy of the trained model. In \cite{07} and \cite{08}, the authors study non-orthogonal multiple access (NOMA) based FL, which allows multiple edge devices to share channels at the same time and frequency, greatly improving spectral efficiency. However, In order to separate signals from various devices, NOMA needs to perform complex successive interference cancellation (SIC) on the receiver. In \cite{09} and \cite{10}, the authors proposed a hierarchical federated learning framework by deploying some edge servers close to devices. In this framework, edge devices perform multiple rounds of edge aggregation on nearby edge servers, and then collect these edge aggregation results to a higher level global server. \cite{11} and \cite{12} utilize over-the-air computation (AirComp) to design a FL framework that integrates communication and computation. AirComp exploits the superposition property of wireless channels to directly aggregate all signals within the same channel resource block. Therefore, the scale of FL is no longer limited by bandwidth resources. In addition, the computation of gradient aggregation (partial calculations performed by FL) is completed in the process of signal transmission, thus greatly improving the efficiency of FL. However, the distortion of the aggregated signal is affected by the device with the worst channel condition (straggler), and the straggler will cause a decline of the training quality in FL. In order to improve the quality of the aggregated signal in AirComp, \cite{13}, \cite{14} and \cite{15} have studied the device scheduling problem. Their goal is to achieve minimal distortion of the received signal while maximizing the number of devices involved in FL. However, these methods will discard some devices with poor channel conditions and waste the data resources of them. Relay communication is a commonly used technique for expanding communication scale. Compared to technologies such as reconfigurable intelligent surface (RIS) and massive multiple-input multiple-output (mMIMO), the advantage of relay communication lies in its relative simplicity. The relay node amplifies the signal to compensate for signal attenuation, without the need for complex multi antenna configurations or reflective surface technology. There are few works that combine relay communication with AirComp-based FL system. The authors in \cite{16} utilize relay to help devices to complete local gradient transmission. However, all devices in this work communicate with  parameter server (PS) through the relay node, which is unnecessary for some devices with perfect channel conditions. In addition, existing works assume that the available power of the device in each communication round is constant, which does not take into account the power heterogeneity of each device.

In this paper, we utilize relay technique to implement AirComp-based FL. Specially, We formulate a non-convex optimization problem under the constraints of power consumption and signal distortion, with the goal of maximizing the set of devices involved in signal aggregation. At the same time, the communication modes of each device have been distinguished. Furthermore, we have dynamically set the available power of each device to make it easier to achieve large-scale FL training in the later communication rounds. To solve this non-convex optimization problem, we propose a device scheduling algorithm, which transforms the non-convex optimization problem into several sparse optimization sub-problems.

The remainder of this article is organized as follows. Section II introduces the system model of FL and AirComp, and problem formulation. Section III shows our relay scheduling algorithm and solves the non-convex problem. The performances of the proposed approach are illustrated in Section IV, and conclusions are given in Section V.

\section{System Model and Problem Formulation}
In our proposed relay-assisted FL system, there is a parameter server (PS)  and a relay node, as well as $K$ edge devices denoted by ${\cal K} = \left\{ {1,2,...,K} \right\}$. Both the PS and edge devices are configured with a single antenna.
These edge devices collaborate to train a machine learning model $\bm{\omega}  \in {\mathbb{R}^d}$ under the coordination of the PS. In addition, each edge device $k \in {\cal K}$ has its own dataset represented as ${{\cal D}_k}$, and the data samples belonging to device $k$ are denoted by $({\emph{\textbf{x}}_k},{y_k}) \in {{\cal D}_k}$, where ${\emph{\textbf{x}}_k}$ and ${y_k}$ represent sample features and sample labels, respectively.

Edge devices are divided into direct transmission devices and cooperative transmission devices (assisted by the relay node). There are two phases of data transmission from edge devices to the PS: (1) The relay node receives aggregated signals from cooperative transmission devices. (2)  The PS receives aggregated signals from the relay node and direct transmission devices.
\begin{figure}[h]
\centering
    \includegraphics[scale=0.45]{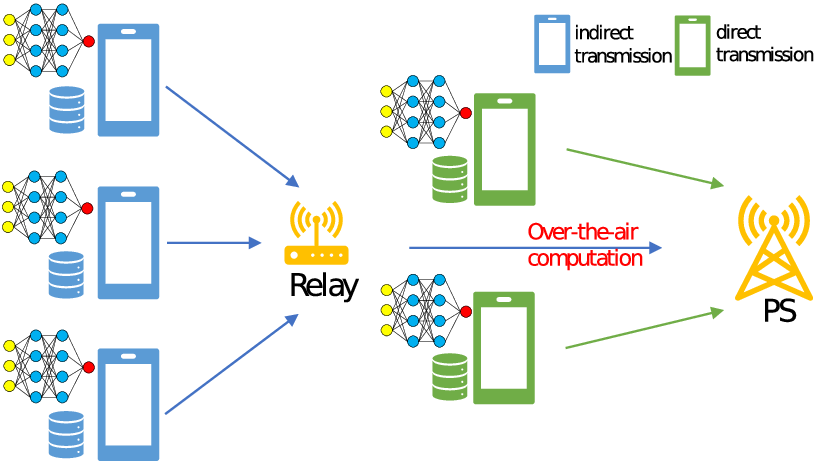}
    \caption{Federated Learning with Relay Communication}
     \vspace{-4mm}
\end{figure}

\subsection{Federated learning}

FL is an iterative training process of the following three steps: (1) The PS assigns the global model to some edge devices. (2) These edge devices use the received global model and their own data for training to obtain a local gradient ${{\textbf{g}}_{k,t}} \in {\mathbb{R}^d}$. (3) the PS aggregates local gradients obtained from these edge devices to update the global model $\bm{\omega}$. Each iteration is called a communication round. The iteration is terminated when the model converges or reaches a predetermined number of communication rounds \cite{03}.

The goal of FL is to minimize the global loss function, which is given by:
\vspace{-2mm}
\begin{equation}
F\left( \bm{\omega}  \right) = \frac{1}{{|{\cal D}|}}\sum\limits_{k = 1}^K {\sum\limits_{{\emph{\textbf{x}}_k} \in {{\cal D}_k}}^{} {f({\emph{\textbf{x}}_k},{y_k};\bm{\omega} )} },
\end{equation}
where $f({\emph{\textbf{x}}_k},{y_k};\bm{\omega} )$ is the sample loss function, which measures the difference between the model's prediction  label $\hat y = \bm{\omega} {\emph{\textbf{x}}_k} + b$ and the actual target ${y_k}$.
$\left| {\cal D} \right| = \sum\limits_{k = 1}^K {\left| {{{\cal D}_k}} \right|} $, where $|{{\cal D}_k}|$ is the volume of ${{\cal D}_k}$. The loss function $F\left( \bm{\omega}  \right)$ can reflect the estimation performance of the model parameter $\bm{\omega}$ on the all samples.

Furthermore, in local training, the local gradient is usually obtained by stochastic gradient descent (SGD) algorithm, which is:
\vspace{-2mm}
\begin{equation}
{\textbf{g}_{k,t}} = \frac{1}{{{L_b}}}\sum\limits_{\emph{\textbf{x}} \in {L_{k,t}}}^{} \nabla  f\left( {\emph{\textbf{x}},y;{\bm{\omega} ^{t - 1}}} \right),
\end{equation}
where the subscripts $t$ and $t-1$ denote the $t$-th communication round and the previous communication round, respectively. $\nabla f\left( {\emph{\textbf{x}},y;{\bm{\omega} ^{t - 1}}} \right)$ is the derivative of the loss function $f\left( {\emph{\textbf{x}},y;{\bm{\omega} ^{t - 1}}} \right)$. ${L_{k,t}} \in {{\cal D}_k}$ is a subset of the local dataset that is randomly and uniformly chosen, and ${L_b} = \left| {{L_{k,t}}} \right|$ is the size of the mini-batch.

We assume that the power consumed by using the SGD algorithm for one data sample is $\nu $, so the computational power consumption of device $k$ in a communication round is
\begin{equation}
E_k^{Comp} = \nu {L_b}.
\end{equation}


After local training, the PS aggregates the local gradients and updates the global model by:
\vspace{-2mm}
\begin{equation}
{\bm{\omega} ^t} = {\bm{\omega} ^{t - 1}} - {\eta _t}\frac{{\sum\limits_{k \in {{\cal S}_t}}^{} {{\textbf{g}_{k,t}}} }}{{\left| {{{\cal S}_t}} \right|}},
\end{equation}
where ${\eta _t}$ represents the learning rate. ${{\cal S}_t}$ and $\left| {{{\cal S}_t}} \right|$ represent the set and the number of the scheduled devices in communication round $t$, respectively.

\subsection{Over-the-air computation aggregation}
In the process of gradient aggregation, edge devices can communicate with the PS directly or indirectly. We use sets ${{\cal S}_d}$ (direct) and ${{\cal S}_i}$ (indirect) to represent the sets of devices in both modes. Local gradients are transmitted using over-the-air computation (AirComp). Specifically, AirComp uses the superposition characteristics of the channel to realize gradient signal aggregation \cite{11}.

For simplicity, we omit the subscript $t$ in the following presentation. To facilitate power control, we normalize the gradient signal to be sent using unit variance, e.g. ${\textbf{s}_k}: = {\textbf{g}_k}$, ${\textbf{s}_k} \in {\mathbb{C}^d}$ and $E({\textbf{s}_k}\textbf{s}_k^H) = \textbf{I}$. After receiving the normalized aggregated gradient, it is necessary to obtain the true gradient information through de-normalization to update the global model $\bm{\omega} $ \cite{16}.

In communication round $t$, the PS receives the aggregated gradient signal from devices $k \in {{\cal S}_d}$ and the relay node, which is given by
\vspace{-2mm}
\begin{equation}
\textbf{s} = \sum\limits_{k \in {{\cal S}_d}}^{} {d_k^{ - \frac{\alpha }{2}}} {h_k}{p_k^d}{\textbf{s}_k} + {f_n}b\textbf{z} + {\textbf{\emph{n}}_1}
\end{equation}
where ${h_k \in {\mathbb{C}}}$ and ${d_k \in {\mathbb{C}}}$ represent the channel coefficient and distance between device $k$ and the PS, respectively. $\alpha $ represents the path loss exponent. ${p_k^d} \in {\mathbb{C}}$ is the transmit scalar of those direct transmission devices. ${\emph{\textbf{n}}_1} \sim {\cal C}{\cal N}(0,{\sigma ^2})$ is additive white Gaussian noise. ${f_n}$ is the channel parameter (channel coefficient and channel fading) between the relay node and the PS. $b$ is the amplify scalar of the relay node. $\textbf{z}$ represents the aggregated signal of device $k \in {{\cal S}_i}$ at the relay node, which is
\begin{equation}
\textbf{z} = \sum\limits_{k \in {{\cal S}_i}}^{} {r_k^{ - \frac{\alpha }{2}}} {j_k}{p_k^i}{\textbf{s}_k} + {\textbf{\emph{n}}_2}{\rm{  = }}\sum\limits_{k \in {{\cal S}_i}}^{} {{\lambda _2}} {\textbf{s}_k} + {\textbf{\emph{n}}_2},
\end{equation}
where ${j_k \in {\mathbb{C}}}$ and ${r_k \in {\mathbb{C}}}$ represent the channel coefficient and distance between the edge device and the relay node, respectively.  ${p_k^i} \in {\mathbb{C}}$ is the transmit scalar of device $k$ in cooperative transmission. ${\emph{\textbf{n}}_2} \sim {\cal C}{\cal N}(0,{\sigma ^2})$ is the additive white Gaussian noise between the device and the relay node.

In order to collect amplitude-aligned aggregated signal at PS, let ${p_k^d} = \frac{{{\lambda _1}}}{{d_k^{ - \frac{\alpha }{2}}{h_k}}}$ and ${p_k^i} = \frac{{{\lambda _2}}}{{r_k^{ - \frac{\alpha }{2}}{j_k}}}$, where $\lambda _1$ and $\lambda _2$ are the power scalar that can affect the received SNR in PS and relay node, respectively. In addition, Let $b = \frac{{{\lambda _1}}}{{{\lambda _2}{f_n}}}{\rm{    }}$, so formula (5) can be simplified to
\begin{equation}
\begin{array}{c}
\textbf{s} = \sum\limits_{k \in {{\cal S}_d} \cup {{\cal S}_i}}^{} {{\lambda _1}} {\textbf{s}_k} + \frac{{{\lambda _1}}}{{{\lambda _2}}}{\textbf{\emph{n}}_1} + {\textbf{\emph{n}}_2}.
\end{array}
\end{equation}

By scaling $\textbf{s}$, we can obtain the aggregated signal
\begin{equation}
\bm{\hat s} = \sum\limits_{k \in {S_d} \cup {S_i}} {{{\bf{s}}_k}}  + \frac{1}{{{\lambda _2}}}{{\bf{\emph{\textbf{n}}}}_1} + \frac{1}{{{\lambda _1}}}{{\bf{\emph{\textbf{n}}}}_2}.
\end{equation}
After that, $\bm{\hat s}$ is de-normalized and substituted into (4) to update the global model.

The communication power consumption of device $k$ for direct transmission is
\begin{equation}
E_k^{tr} = {\left| {{p_k^d}{\textbf{s}_k}} \right|^2} = \frac{{\lambda _1^2}}{{d_k^{ - \alpha }h_k^2}},{k} \in {{\cal S}_d}.
\end{equation}
Similarly, the communication power consumption of device $k$ for cooperative transmission is
\begin{equation}
E_k^{tr} = {\left| {{p_k^i}{\textbf{s}_k}} \right|^2} = \frac{{\lambda _2^2}}{{r_{k}^{ - \alpha }j_{k}^2}},{k} \in {{\cal S}_i}.
\end{equation}
The power consumption of the relay node is
\begin{equation}
{E_{relay}} = {\left| {b\textbf{z}} \right|^2} = \frac{{\left| {{{\cal S}_i}} \right|}}{{f_n^2}}\lambda _1^2 + \frac{1}{{f_n^2}}\frac{{\lambda _1^2}}{{\lambda _2^2}}{\sigma ^2}.
\end{equation}

The above system model is shown in Fig.1.

It is easy to see that the actual aggregated signal suffers from distortions caused by noise and fading. We use the mean squared error (MSE) to measure the magnitude of the distortion, which is defined as:
\begin{equation}
MSE =E \left( {{{\left| {{\bm{\hat s}} - \sum\limits_{k \in {S_d} \cup {S_i}} {{\textbf{s}_k}} } \right|}^2}} \right) = \left(\frac{1}{{\lambda _2^2}} + \frac{1}{{\lambda _1^2}}\right){\sigma ^2}.
\end{equation}

\subsection{Problem formulation}
In FL, increasing the number of devices means that more data is available for model updates, resulting in high-accuracy models. However, due to the power consumption and the MSE constraint, the number of scheduled devices cannot be increased without limitation. Therefore, the goal of our optimization problem is to maximize the number of devices involved in training while meeting power consumption and MSE constraints. In communication round $t$, the optimization problem to be solved is shown as follows:
\begin{equation}
\begin{array}{l}
\mathop {\max }\limits_{{\lambda _1},{\lambda _2},{{\cal S}_d},{{\cal S}_i}} \left| {{{\cal S}_d}} \right| + \left| {{{\cal S}_i}} \right|\\
\nu {L_b} + \frac{{\lambda _1^2}}{{d_k^{ - \alpha }h_k^2}} \le \frac{{{P_k}{\rm{ -  }}\sum\limits_{\tau  = 0}^{t - 1} {{E_{k,\tau }}} }}{{T - t}}{\rm{  ,  }}k \in {{\cal S}_d}\\
\nu {L_b} + \frac{{\lambda _2^2}}{{r_k^{ - \alpha }j_k^2}} \le \frac{{{P_k}{\rm{ -  }}\sum\limits_{\tau  = 0}^{t - 1} {{E_{k,\tau }}} }}{{T - t}}{\rm{  ,  }}k \in {{\cal S}_i}\\
\frac{{\left| {{{\cal S}_i}} \right|}}{{f_n^2}}\lambda _1^2 + \frac{1}{{f_n^2}}\frac{{\lambda _1^2}}{{\lambda _2^2}}{\sigma ^2} \le {P_{\rm{relay}}}\\
\frac{1}{{\lambda _1^2}}{\sigma ^2} + \frac{1}{{\lambda _2^2}}{\sigma ^2} \le \gamma
\end{array},
\end{equation}
where ${\sum\limits_{\tau  = 0}^{t - 1} {{E_{k,\tau }}} }$ represents the communication and computation power consumptions of device $k$ before the $t$-th communication round. Note that many studies set the available power for each communication round to a fixed value, which results in the unused power of the current round not being used in the future. ${{P_k}}$ is the total available power of device $k$ for FL. ${P_{\rm{relay}}}$ and $\gamma $ denote the relay power consumption constraint and the MSE constraint at the PS, respectively.

Unfortunately, it is difficult to solve optimization problem (13) due to the combinatorial nature with the objective function and the non-convex constraints. According to \cite{11}, we can transform the original optimization problem into a sparse optimization problem. Specifically, We introduce a binary variable ${a_k} \in \{ 0,1\} $, where ${a_k}$=1 indicates that device $k$ is subject to the power constraint of cooperative transmission, and otherwise is subject to direct transmission. Then the transformed sparse optimization problem is show as follows:
\begin{equation}
\begin{array}{l}
\mathop {\min }\limits_{{\lambda _1},{\lambda _2},\emph{\textbf{a}},\emph{\textbf{m}}} {\left\| \emph{\textbf{m}} \right\|_0}\\
\nu {L_b} + \left( {1 - {a_k}} \right)\frac{{\lambda _1^2}}{{d_{k}^{ - \alpha }h_{k}^2}}\\
 + {a_k}\frac{{\lambda _2^2}}{{r_{k}^{ - \alpha }j_{k}^2}} - \frac{{{P_k}{\rm{ -  }}\sum\limits_{\tau  = 0}^{t - 1} {{E_{k,\tau}}} }}{{T - t}}{\rm{  }} \le {m_k}{\rm{,  }}\forall k = 1,...,K\\
\frac{{\left| {{{\cal S}_i}} \right|}}{{f_n^2}}\lambda _1^2 + \frac{1}{{f_n^2}}\frac{{\lambda _1^2}}{{\lambda _2^2}}{\sigma ^2} \le {P_{\rm{relay}}}\\
\frac{1}{{\lambda _1^2}}{\sigma ^2} + \frac{1}{{\lambda _2^2}}{\sigma ^2} \le \gamma \\
{a_k} \in \left\{ {0,1} \right\},{\rm{  }}\forall k = 1,...,K
\end{array},
\end{equation}
where $\emph{\textbf{m}} = ({m_1},...,{m_K})$ is a sparse variable. $m_k = 0$ means that device $k$ satisfy the power constraint, that is, device $k$ can participate in the current round of training. ${\left\| \emph{\textbf{m}} \right\|_0}$ denotes the number of nonzero elements in $\emph{\textbf{m}}$, namely the zero norm. $\emph{\textbf{a}} = ({a_1},...,{a_K}).$
In the following sections, we solve optimization problem (14) based on the proposed device scheduling algorithm.

\section{Device Scheduling and Power Allocation}
Although the above optimization problem is transformed into a sparse optimization problem, it is still intractable due to the non-convex objective function and constraint. To overcome these difficulties, we propose a device scheduling algorithm as follows.

\subsection{Device scheduling algorithm and power allocation}
we first define the priority of each device to use the cooperative transmission mode as
\begin{equation}
{\psi _k} = (\frac{{{P_k} - \sum\limits_{\tau  = 0}^{t - 1} {{E_{k,\tau }}} }}{{T - t}} - \nu {L_b})r_k^{ - \alpha }j_k^2,\forall k = 1,...,K,
\end{equation}
which takes into account both the available power and the channel conditions (large ${\psi _k}$ have higher priority).
Sort ${\psi _k},{\text{ }}\forall k = 1,...,K$ from largest to smallest, and let $\psi _k^i$ denote the $i$-th ranked ${\psi _k}$.

Starting from $\textbf{\emph{a}} = \textbf{0}$, according to the priority $\psi _k^i$, let the corresponding ${a_k} = 1$. Each time the value of $\emph{\textbf{a}}$ is updated, a device is added to ${{\cal S}_i}$  and the value of $\left| {{{\cal S}_i}} \right|$ can be obtained. At the same time, let ${\lambda _2} = \sqrt {\psi _k^i}$.
In the subsequent performance analysis, the reason for taking this value are explained. It can be observed that the value of ${\lambda _2}$ is gradually decreasing with $\psi _k^i$, which ensures that all devices $k \in {\mathcal{S}_i}$ satisfy the power consumption constraint. When $\left| {{\mathcal{S}_i}} \right|$ and ${\lambda _2}$ is fixed, obtain the following sub-optimization problem
\begin{equation}
\begin{array}{l}
\mathop {\min }\limits_{{\lambda _1},\emph{\textbf{m}}} {\left\| \emph{\textbf{m}} \right\|_0}\\
v{L_b} + \frac{{\lambda _1^2}}{{d_k^{ - \alpha }h_k^2}} - \frac{{{P_k} - \sum\limits_{\tau  = 0}^{t - 1} {{E_{k,\tau }}} }}{{T - t}} \le {m_k},\forall k \in \{ k|{a_k} = 0\} \\
\lambda _1^2 \le \frac{{{P_{\rm{relay}}}f_n^2\lambda _2^2}}{{\left| {{{\cal S}_i}} \right|\lambda _2^2 + {\sigma ^2}}}\\
\lambda _1^2 \ge \frac{{{\sigma ^2}\lambda _2^2}}{{\gamma \lambda _2^2 - {\sigma ^2}}}
\end{array}
\end{equation}
after shifting terms and common partitioning.
It can be observed that the more devices $k \in \{ k|{a_k} = 0\}$ satisfy constraint $\nu {L_b} + \frac{{\lambda _1^2}}{{d_k^{ - \alpha }h_k^2}} - \frac{{{P_k}-\sum\limits_{\tau  = 0}^{t - 1} {{E_{k,\tau }}} }}{{T - t}} \leqslant 0$, the smaller the value of ${\left\| {\mathbf{\textbf{\emph{m}}}} \right\|_0}$, that is, $\lambda _1$ takes the minimum value. Therefore, when $\frac{{{\sigma ^2}\lambda _2^2}}{{\gamma \lambda _2^2 - {\sigma ^2}}} \leqslant \frac{{{P_{\rm{relay}}}f_n^2\lambda _2^2}}{{\left| {{\mathcal{S}_i}} \right|\lambda _2^2 + {\sigma ^2}}}$ (${P_{\rm{relay}}}f_n^2 \ge \frac{{{\sigma ^2}(\left| {{S_i}} \right|\lambda _2^2 + {\sigma ^2})}}{{\gamma \lambda _2^2 - {\sigma ^2}}}$), the optimal value of ${\lambda _1}$ is given by $\lambda _1^2 \geqslant \frac{{{\sigma ^2}\lambda _2^2}}{{\gamma \lambda _2^2 - {\sigma ^2}}}$, i.e. $\lambda _1^ *  = \sqrt {\frac{{{\sigma ^2}\lambda _2^2}}{{\gamma \lambda _2^2 - {\sigma ^2}}}} $.

Proposition 1: When ${P_{\rm{relay}}}f_n^2 \ge \frac{{{\sigma ^2}(\left| {{S_i}} \right|\lambda _2^2 + {\sigma ^2})}}{{\gamma \lambda _2^2 - {\sigma ^2}}}$ is not satisfied for the first time, there is no need to update the value of $\textbf{\emph{a}}$.

Proof: Inequality ${P_{\rm{relay}}}f_n^2 \ge \frac{{{\sigma ^2}(\left| {{S_i}} \right|\lambda _2^2 + {\sigma ^2})}}{{\gamma \lambda _2^2 - {\sigma ^2}}}$ can be transformed to ${P_{\rm{relay}}}f_n^2 \ge \frac{{{\sigma ^2}\left| {{S_i}} \right|{\rm{ + }}{\sigma ^{\rm{4}}}{\rm{/}}\lambda _{\rm{2}}^{\rm{2}}}}{{\gamma  - {\sigma ^2}{\rm{/}}\lambda _{\rm{2}}^{\rm{2}}}}$.
 As $\emph{\textbf{a}}$ updates, the value of ${\lambda _2}$ decreases and the value of $\left| {{{\cal S}_i}} \right|$ increases. Therefore, all subsequent values of ${\lambda _2}$ and $\left| {{{\cal S}_i}} \right|$ cannot satisfy the above inequality.

In addition, when ${\lambda _2} < \sqrt {\frac{{{\sigma ^2}}}{\gamma }} $, all subsequent values of ${\lambda _2}$ cannot satisfy the MSE constraint of the AirComp, so the above repeated process can be terminated to reduce the computational complexity.

We use ${\left\| \textbf{\emph{m}} \right\|_{0,i}}$ to represent the solution when adding the $i$-th priority device to ${{\cal S}_i}$. The minimum value of ${\left\| \textbf{\emph{m}} \right\|_0}$ is the optimal solution of the current communication round.

When no edge device in ${{\cal S}_i}$, the optimization problem becomes
\begin{equation}
\begin{gathered}
  \mathop {\min }\limits_{{\lambda _1},\textbf{\emph{m}}} {\left\| \textbf{\emph{m}} \right\|_0} \hfill \\
  v{L_b} + \frac{{\lambda _1^2}}{{d_k^{ - \alpha }h_k^2}} - \frac{{{P_k} - \sum\limits_{\tau  = 0}^{t - 1} {{E_{k,\tau }}} }}{{T - t}} \leqslant {m_k},\forall k = 1,...,K \hfill \\
  \frac{1}{{\lambda _1^2}}{\sigma ^2} \leqslant \gamma.  \hfill \\
\end{gathered}
\end{equation}
It can be seen from constraint $\frac{1}{{\lambda _1^2}}{\sigma ^2} \leqslant \gamma $ that $\lambda _1^ *  = \sqrt {\frac{{{\sigma ^2}}}{\gamma }} $. Given the number of devices by $\nu {L_b} + \frac{{{{\left( {\lambda _1^ * } \right)}^2}}}{{d_k^{ - \alpha }h_k^2}} - \frac{{{P_k}-\sum\limits_{\tau  = 0}^{t - 1} {{E_{k,\tau }}} }}{{T - t}} \leqslant 0$, we can obtain the value of ${\left\| {\textbf{\emph{m}}} \right\|_{0,0}}$.

\begin{algorithm}[!h]
\caption{Device scheduling algorithm}
\quad intial $\emph{\textbf{a}} = \textbf{0}$,

\quad ${\psi _k} = (\frac{{{P_k} - \sum\limits_{\tau  = 0}^{t - 1} {{E_{k,\tau }}} }}{{T - t}} - \nu {L_b})r_k^{ - \alpha }j_k^2,\forall k = 1,...,K$.

\quad sort $\psi _k$ as $\psi _k^1,\psi _k^2,...,\psi _k^K$.

\quad \textbf{for} $i = 1,2,...K$ \textbf{do}

\quad \quad ${\lambda _2} = \sqrt {\psi _k^i} $.

\quad \quad \textbf{if} ${P_{\rm{relay}}}f_n^2 \ge \frac{{{\sigma ^2}(\left| {{S_i}} \right|\lambda _2^2 + {\sigma ^2})}}{{\gamma \lambda _2^2 - {\sigma ^2}}}$ \textbf{and} ${\lambda _2} > \sqrt {\frac{{{\sigma ^2}}}{\gamma }} $

\qquad \quad $\lambda _1^ *  = \sqrt {\frac{{{\sigma ^2}\lambda _2^2}}{{\gamma \lambda _2^2 - {\sigma ^2}}}} $, $\left| {{\mathcal{S}_d}} \right| = 0$

\quad \quad \quad\textbf{for} $k = 1,2,...K$ \textbf{do}

\quad \quad \quad \quad \textbf{if} ${a_k} = 0$ \textbf{and} $\nu {L_b} + \frac{{(\lambda _1^ *)^2}}{{d_k^{ - \alpha }h_k^2}} - \frac{{{P_k}-\sum\limits_{\tau  = 0}^{t - 1} {{E_{k,\tau }}} }}{{T - t}} \leqslant 0$

\quad \quad \quad \quad \quad $\left| {{\mathcal{S}_d}} \right| = \left| {{\mathcal{S}_d}} \right| + 1$

\quad \quad \quad \quad \textbf{end if}

\quad \quad \quad \textbf{end for}

\quad \quad \quad ${\left\| \textbf{\emph{m}} \right\|_{0,i}} = K - i - \left| {{\mathcal{S}_d}} \right|$

\quad \quad \textbf{else}

\qquad \quad return

\quad \quad \textbf{end if}

\quad \textbf{end for}

\quad ${\left\| \textbf{\emph{m}} \right\|_0} = \min \{ {\left\| \textbf{\emph{m}} \right\|_{0,0}},...,{\left\| \textbf{\emph{m}} \right\|_{0,K}}\} $

\end{algorithm}

\subsection{Performance analysis}
We prove in this section that after determining the value of $\left| {{\mathcal{S}_i}} \right|$, the designed algorithm yields the maximum value of $\left| {{\mathcal{S}_d}} \right|$. Suppose that the device with the $i$-th priority is added to the device set ${\mathcal{S}_i}$, where $\left| {{\mathcal{S}_i}} \right| = i$. In this time, when $\sqrt {\psi ^i} \geqslant {\lambda _2} > \sqrt {\psi ^{i + 1}}$, only devices in ${\mathcal{S}_i}$ meet the conditions for cooperative transmission. On the other hand, $\lambda _1^ *  = \sqrt {\frac{{{\sigma ^2}\lambda _2^2}}{{\gamma \lambda _2^2 - {\sigma ^2}}}}  = \sqrt {\frac{{{\sigma ^2}}}{{\gamma  - {\sigma ^2} / \lambda _2^2}}} $. When ${\lambda _2}$ takes the maximum value $ \sqrt {\psi ^i} $, $\lambda _1^ * $ takes the minimum value. Thus, $\left| {{\mathcal{S}_d}} \right|$ reaches the maximum value. We obtain all values of $\left| {{\mathcal{S}_i}} \right|$ and found the corresponding maximum number of $\left| {{\mathcal{S}_i}} \right| + \left| {{\mathcal{S}_d}} \right|$ by the proposed method.

\subsection{Computational complexity}
When calculating all the priorities ${\psi _k}$, the complexity is ${\rm O}({\rm \emph{K}})$. The complexity of sorting the priorities is bounded by ${\rm O}({{\rm \emph{K}}^2})$, e.g. quick sort and insertion sort. In addition, the complexity of obtaining the minimum of ${\left\| \textbf{\emph{m}} \right\|_0}$ are determined by two factors: The number of times for adding devices to ${\mathcal{S}_i}$ (at most \emph{K} times); the complexity to get each ${\left\| \textbf{\emph{m}} \right\|_{0,i}}$ (calculate ${\lambda _1}$, verify ${P_{\rm{relay}}}f_n^2 \geqslant \frac{{{\sigma ^2}(\left| {{\mathcal{S}_i}} \right|\lambda _2^2 + {\sigma ^2})}}{{\gamma \lambda _2^2 - {\sigma ^2}}}$ and compare each device power consumption constraint) at most ${\rm O}({\rm \emph{K}} + 2)$. The complexity of verifying ${\lambda _2} < \sqrt {\frac{{{\sigma ^2}}}{\gamma }} $ is at most ${\rm O}({\rm \emph{K}})$. The maximum complexity of comparing all ${\left\| \textbf{\emph{m}} \right\|_{0,i}}$ is \emph{K}. As a result, the overall complexity of the proposed device scheduling algorithm is upper bounded by ${\rm O}({\rm \emph{K}}({\rm \emph{K}} + 2) +\rm \emph{K}+ {{\rm \emph{K}}^2} + {\rm \emph{K}}+ {\rm \emph{K}}) = {\rm O}(2{{\rm \emph{K}}^2} + 5{\rm \emph{K}})$, implying that the proposed problem can be solved in polynomial time.



\section{Simulation Results}
In this section, we conduct experimental simulations to evaluate the performance of FL based on our proposed device scheduling algorithm. We use FL to train a ResNet-18 model with the goal of classifying the CIFAR-10 image dataset. ResNet18 is a classical deep convolutional neural network model, which contains 18 convolutional layers. The CIFAR-10 dataset consists of 60,000 color images, which are evenly divided into 10 classes. In this experiment, we set up a cellular system with the radius of 100 meters. The location of the PS is at (0,0) and the relay node is at (50,50). It is assumed that there are 20 mobile devices randomly distributed in the first quadrant of the cellular system. The channel coefficients ${h_k}$ and ${j_k}$ follow Rayleigh fading (${h_k},{j_k} \sim {\cal C}{\cal N}\left( {0,1} \right)$) and are independently and identically distributed (i.i.d.) with respect to the parameters $k$ and $t$. Assumed that the path loss exponent $\alpha  = 3$. The learning rate ${\eta _t}$ is set to 0.001 and the mini-batch size ${L_b}$ is 32. We assume that all devices have the same dataset size and use the same mini-batch size during training, so the power consumption of local computation is ignored, i.e., $E_k^{Comp} \to 0$. The other parameters are set as:${P_k} = 5W$, ${P_{\rm{relay}}} = 1W$, $\gamma  = 5dB$, ${\sigma ^2} = {10^{ - 6}}W$, terminate round $T = 100$.

To show the superiority of our proposed scheme, we consider the following benchmark schemes:

\begin{itemize}

  \item Transmission without relay node: In this scheme, all devices can only communicate with the PS by direct transmission. The available power of all devices in each communication round is fixed to ${P_k}/T$.
  \item The relay-assisted scheme in \cite{16}: In this scheme, all devices participate the transmission of gradients via the assistance of a relay node. The available power of the device for FL is a fixed value ${P_k}/T$.
  \item Ideal relay channel: This scheme is similar to the second one except that the channel coefficient between the relay node and the PS is set to 1.  This is an ideal benchmark.
\end{itemize}
\begin{figure}[h]
\vspace{-4.15mm}
\centering
    \includegraphics[scale=0.45]{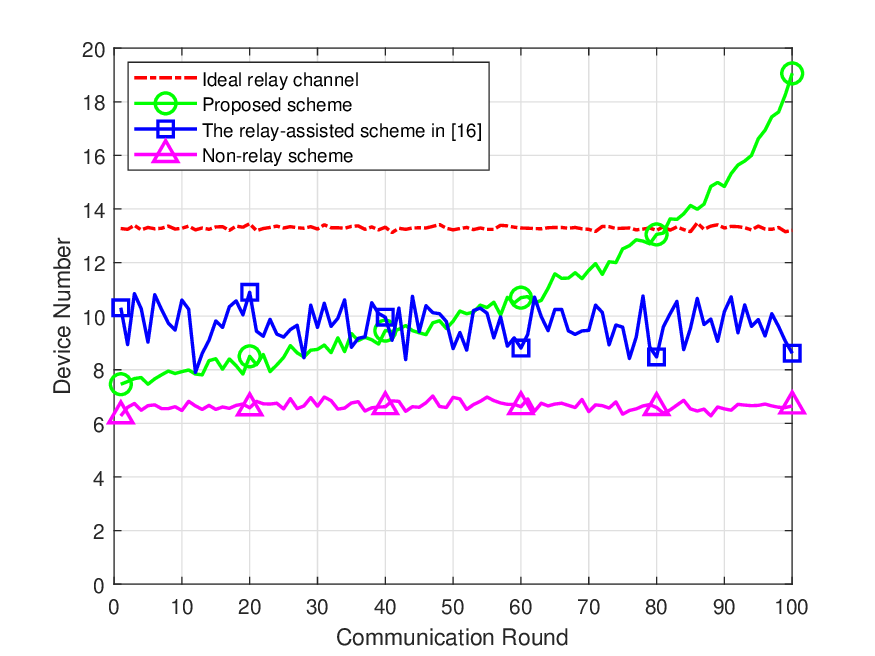}
    \caption{Number of devices for different schemes}
    \vspace{-4.15mm}
\end{figure}

As shown in Fig. 2, We compared the number of devices scheduled for different schemes. It can be observed that compared with the scheme in \cite{16}, the proposed scheme can gradually increase the number of devices involved in training. The reason for this phenomenon is the dynamic power setting of the device scheduling algorithm, which can make the device that has not participated in the training to store power, so that the device can participate in the training more easily during subsequent communication rounds. In addition, when the channel conditions of the relay node are poor, the devices in the proposed scheme can choose the direct transmission method, so the robustness is stronger. It can be seen that the relay scheme in [16] fluctuates significantly under changing relay channel conditions. Because edge devices with poor channel conditions do not have the assistance of relay nodes, the performance of non-relay scheme is the worst. The ideal relay channel scheme performs consistently well, but as the number of communication rounds increases to more than 80, our proposed scheme performs better than it.

\begin{figure}[h]
\vspace{-4.15mm}
\centering
    \includegraphics[scale=0.45]{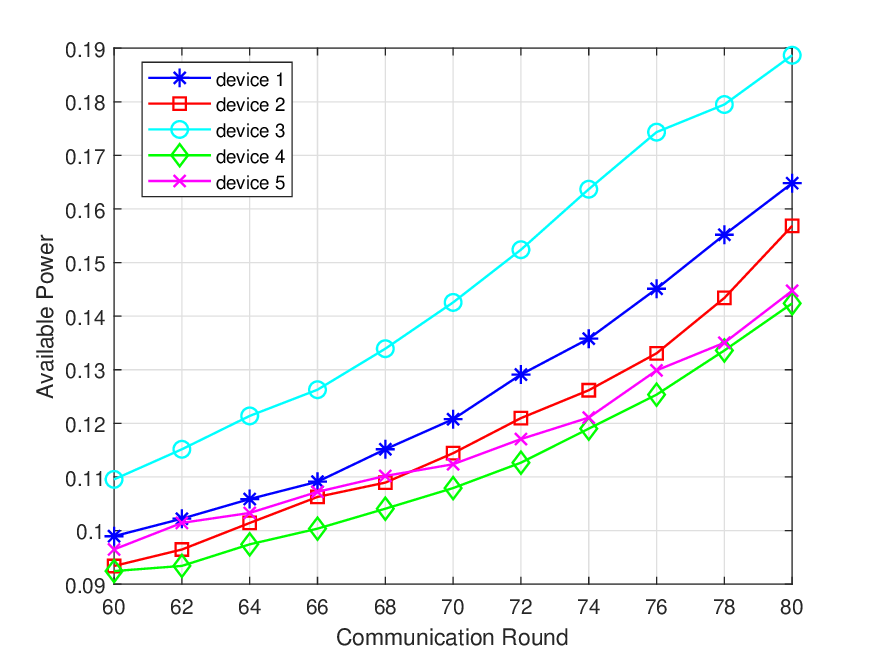}
    \caption{The process of power accumulation}
    \vspace{-4.15mm}
\end{figure}

Fig. 3 illustrates the available power accumulation process of the proposed scheme. We randomly selected five devices and observed their available power changes with the communication rounds. It can be observed that the available power of all devices gradually accumulates as the number of communication rounds increases. This dynamic power setting would make it easier to achieve large-scale training in the later stages of FL. The authors in \cite{02} have demonstrated that increasing the number of edge devices for FL in the later stages is more beneficial for improving model accuracy. This effect is easily accomplished by the proposed algorithm.

\begin{figure} [h]
\vspace{-3mm}
	\centering
	\subfloat[\label{1a}{Test accuracy of $K = 20$}]{
		\includegraphics[scale=0.45]{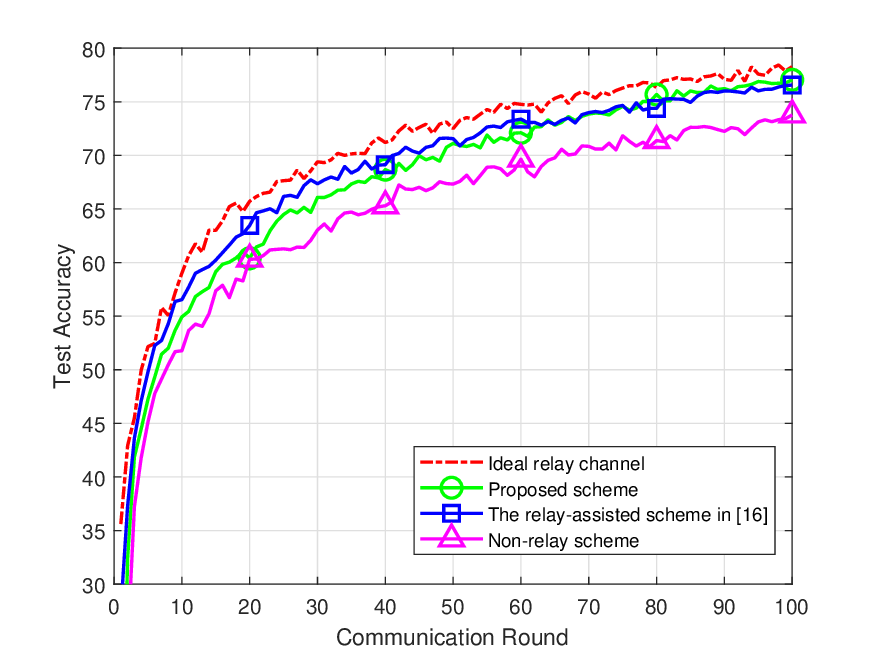}}
	\\
    \vspace{-4mm}
	\subfloat[\label{1b}{Test accuracy for different $K$}]{
		\includegraphics[scale=0.45]{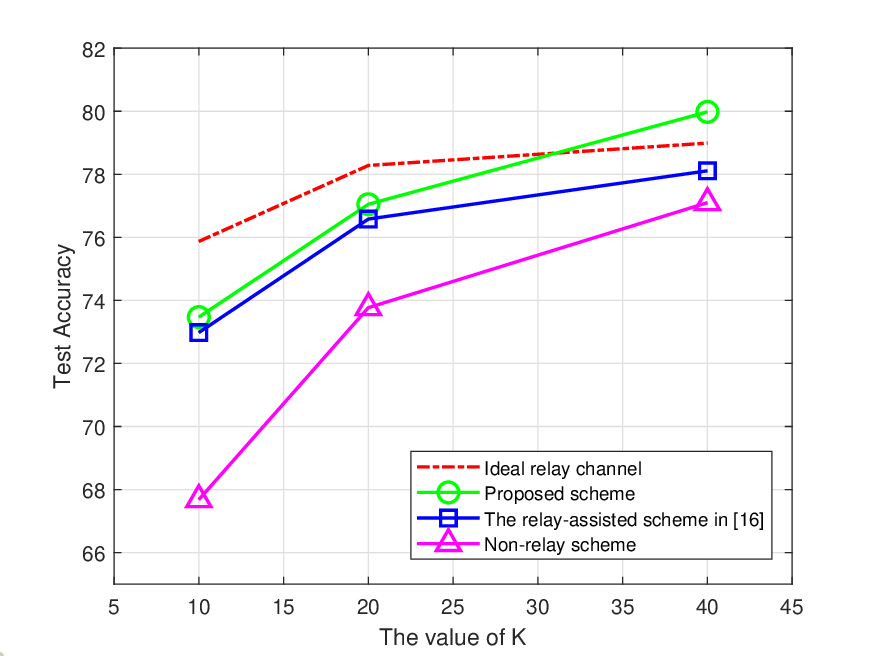} }
	\caption{Test accuracy of different schemes}
	\label{fig2}
    \vspace{-4.15mm}
\end{figure}
Fig. 4 compares the test accuracy of the models trained by different schemes. It can be observed from Fig 4(a) that when the total number of edge devices $K$=20, the test accuracy of the proposed scheme is similar to the relay scheme in \cite{16}. Due to stable and excellent relay channel conditions, the ideal relay channel scheme has the highest accuracy. The number of devices that can participate in FL training in the non-relay scheme has remained at a small level, so its accuracy is significantly smaller than the other three schemes. From Fig. 4(b), it can be seen that as the total number of devices in the system increases, the performance of all solutions improves. This is because more dense devices within the same range enable more devices to transmit their own signals. When $K$ increases, the test accuracy of the proposed scheme performs better than the scheme in \cite{16} and the ideal relay channel scheme. The reason is that as $K$ increases, the relay node gradually reaches the limit of its own load. In this case, the proposed scheme can allocate a part of devices to complete gradient aggregation by direct transmission. Therefore, the proposed scheme has better robustness compared with other schemes.

\section{Conclusions}
In this paper, we propose a novel device scheduling scheme for relay-assisted AirComp in FL, which transforms the formulated non-convex optimization problem into several sparse sub-problems and determines the communication mode of each mobile device. Simulation results show that our proposed scheme outperforms other benchmark schemes. In the future, our work will consider more general scenarios such as multiple relays and multiple antennas.

\bibliography{reference}
\bibliographystyle{IEEEtran}

\end{document}